\documentclass{SCIS2026}

\usepackage[english]{babel}

\usepackage{arydshln}
\usepackage{xcolor}
\usepackage{balance}
\usepackage[T1]{fontenc}
\newtheoremstyle{exampstyle}
{0.0em}
{0.0em}
{}
{1em}
{\bfseries}
{.}
{1em}
{}

\theoremstyle{exampstyle}


\begin{document}
	\ArticleType{RESEARCH PAPER}
	\Year{2021}
	\Month{}
	\Vol{}
	\No{}
	\DOI{}
	\ArtNo{}
	\ReceiveDate{}
	\ReviseDate{}
	\AcceptDate{}
	\OnlineDate{}
	
	\title{SpatialQ: Understanding 3D Gaussian Splatting Scene Quality via Visual-based MLLM}{Title for citation}
	
	\author[1]{Jingxuan SU}{}
    \author[2]{Shenglin WANG}{}
    \author[3]{Tiesong ZHAO}{}
    \author[1]{Ge LI}{}
    \author[1,2]{Wei GAO}{{gaowei262@pku.edu.cn}}
	\AuthorMark{Author A}
	
		\AuthorCitation{Jingxuan Su, Author B, Author C, et al}
	
	

    \address[1]{Guangdong Provincial Key Laboratory of Ultra High Definition Immersive Media Technology,\\ School of Electronic and Computer Engineering, Peking University, Shenzhen 518055, China}
    \address[2]{Peng Cheng Laboratory, Shenzhen, {\rm 518055}, China}
    \address[3]{College of Physics and Information Engineering, Fuzhou University, Fuzhou 350108, China}

	
	\abstract{3D Gaussian Splatting (3DGS) has emerged as an effective representation for novel view synthesis and 3D scene reconstruction, creating an increasing demand for reliable quality assessment. Unlike conventional image quality assessment (IQA), the quality of a 3DGS scene depends not only on the perceptual fidelity of rendered views, but also on scene-level factors such as spatial structure and cross-view consistency. Existing IQA methods are limited by their reliance on 2D perceptual cues, whereas general multimodal large language models (MLLMs) are not designed for stable quality regression and may produce unreliable judgments. To address these limitations, a multimodal quality assessment framework is developed for 3DGS scene understanding. First, a 3D-aware quality representation learning framework is introduced by augmenting a VGGT-based encoder with a dedicated quality head. Multi-view images are encoded into view-specific features and aggregated to capture cross-view consistency, while geometric cues are incorporated through joint modeling of depth and point-cloud-related structural information, enabling the learning of structure-aware quality representations beyond appearance-driven features. Second, a grounded multimodal reasoning mechanism is constructed by jointly feeding original images, depth maps, point cloud renderings, and camera parameters into a Qwen-based MLLM. By explicitly exposing geometric continuity from depth, spatial distribution from point clouds, and viewpoint configuration from camera parameters, the model performs structured reasoning to infer degradation types, such as geometric inconsistency, sparsity, and appearance distortion, together with their severity levels from multimodal evidence. Third, a reasoning-guided refinement strategy is designed to decompose quality prediction into a base score and an adjustment term, where the inferred degradation type and severity are translated into a score correction to compensate for biases in the initial prediction while providing interpretable degradation attribution. Extensive experiments demonstrate that the proposed method achieves effective and robust performance on 3DGS scene quality assessment, while producing degradation reasoning that is well aligned with the underlying scene characteristics.}
	\keywords{3D gaussian splatting, visual quality assessment, large language model, multimodal learning, 3D scene understanding }
	
	\maketitle

\section{Introduction}      

Recent advances in 3D Gaussian Splatting (3DGS) have significantly improved the efficiency and visual fidelity of novel view synthesis and large-scale scene reconstruction~\cite{bao20253d,fei20243d}, enabling a wide range of applications in immersive rendering~\cite{yang2025generalizable, wang2026artificial}, robotics perception~\cite{zhu20243d}, and digital twin systems~\cite{lee2026multimodal}. As 3DGS-based representations are increasingly deployed in practical scenarios, reliable quality assessment becomes essential for model selection, optimization, and failure diagnosis~\cite{wu2024recent}. However, assessing the quality of 3DGS scenes is inherently more challenging than conventional image quality assessment (IQA)~\cite{martin2025gs,wan2026perceptual}. Unlike 2D images, where quality is primarily determined by perceptual distortions within a single view, the quality of a 3DGS scene depends on multiple interrelated factors, including geometric plausibility, spatial structure consistency, cross-view coherence, and the validity of camera configurations~\cite{wu2025survey}. Since these quality attributes are not always directly observable from individual rendered images, existing image-based metrics often fail to provide reliable scene-level evaluation~\cite{xing20253dgs}. Nevertheless, because 3DGS scenes are ultimately perceived through rendered views, IQA methods still provide a natural and practically relevant baseline for comparison. Consequently, developing an effective quality assessment method that can jointly capture perceptual fidelity and underlying spatial degradation in 3DGS scenes remains an open and important problem.

Existing image quality assessment (IQA) methods, including classical metrics as well as learning-based and deep no-reference models, provide a natural starting point for evaluating rendered views of 3DGS scenes~\cite{zhang2025quality}. These methods have achieved strong performance in characterizing perceptual distortions in 2D images, particularly for degradations such as blur, noise, and compression artifacts~\cite{zhang2026towards,zhai2020perceptual}. However, their underlying assumption is that quality can be reliably inferred from perceptual cues within a single image, which does not hold for 3DGS scenes~\cite{wan2026perceptual}. Although the quality of a 3DGS scene is ultimately perceived through rendered views, the underlying distortions originate from the 3D representation itself. As a result, reliable quality assessment requires inferring scene-level structural degradation from multi-view image evidence, rather than relying solely on image-level perceptual cues. This creates an inherent mismatch between the scene-level nature of 3DGS quality and the image-level assumptions underlying conventional IQA models. Consequently, IQA-based approaches are fundamentally limited in capturing spatially induced degradations, such as depth inconsistency, structural misalignment, and cross-view instability, while simple score aggregation across rendered views is insufficient to resolve this issue~\cite{chenfast,yang2024benchmark}.

Beyond image-based methods, point cloud quality assessment (PCQA) approaches have also been developed to evaluate geometric distortions in 3D data~\cite{11456849,wang2023applying,fan2025stochasticity,wang2024zoom}. These methods are effective for static geometry evaluation, but do not explicitly account for view-dependent rendering, radiance variation, or cross-view consistency in 3DGS scenes. Quality assessment for neural rendering, such as NeRF-based representations, has further considered rendering fidelity and multi-view consistency~\cite{martin2023nerf,xing2024explicit}. Nevertheless, these methods are not designed for compressed 3DGS representations, where distortions are closely tied to explicit Gaussian structure and compression mechanisms. Thus, these observations suggest that 3DGS quality assessment requires a different formulation that goes beyond both image-level perceptual modeling and geometry-only quality evaluation, and instead demands explicit understanding of scene-level spatial structure and multi-view coherence.

Multimodal large language models (MLLMs) have demonstrated strong capabilities in visual understanding and cross-modal reasoning, making them a potentially valuable tool for addressing the spatial complexity of 3DGS scenes~\cite{zhang2025large}. In particular, their ability to integrate heterogeneous inputs and perform high-level semantic inference suggests clear potential for identifying geometry-related artifacts, cross-view inconsistencies, and other scene-level degradations~\cite{li2025qinsight}. However, extending MLLMs to 3DGS quality assessment still faces several challenges~\cite{zhao2025reasoning}. Since MLLMs are not explicitly optimized for quantitative quality regression, they often lack the stability required for consistent score prediction. Moreover, without structured guidance, their outputs can be sensitive to prompting and may not align well with perceptual quality criteria. Existing MLLM-based evaluation approaches are also primarily developed for general visual understanding rather than quality assessment, and therefore do not explicitly exploit geometric cues such as depth, point cloud structure, or camera configurations, which are essential for 3DGS scene evaluation~\cite{gemini15,zhang2023liqe}. Consequently, MLLMs are better suited to serve as spatially grounded reasoning modules that complement, rather than replace, data-driven quality prediction. This observation motivates a unified framework that combines stable base quality estimation with multimodal reasoning, enabling both accurate score prediction and interpretable degradation analysis for 3DGS scenes.

\begin{figure*}[t]
\centering
\includegraphics[width=\textwidth]{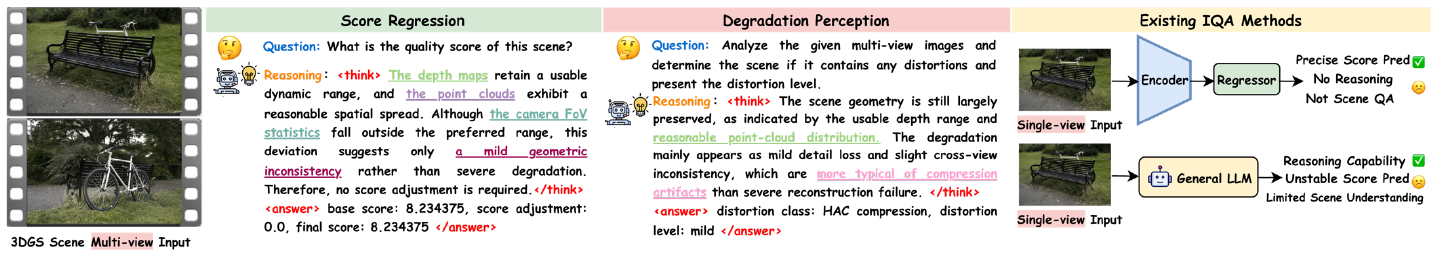}
\caption{Pipeline comparisons between our proposed SpatialQ and existing IQA methods (right). Two examples of our SpatialQ (left) are presented. Additionally, SpatialQ effectively supports quality score adjustment, image degradation perception, and scene spatial reasoning tasks.}
\label{fig1}
\end{figure*}

In this work, these limitations are addressed through a multimodal quality assessment framework that unifies 3D-aware representation learning with reasoning-guided score refinement (as shown in Figure~\ref{fig1}). Specifically, a spatial quality encoder is first introduced to learn scene-level quality representations from multi-view rendered images and to produce a base quality score, which serves as a stable data-driven estimate of the reconstructed scene. On top of this base predictor, a Qwen-based MLLM is employed as a spatially grounded reasoning module rather than a standalone score generator. By jointly analyzing the rendered images together with encoder-derived depth maps, point-cloud visualizations, and camera parameters, the MLLM explicitly diagnoses degradation patterns in terms of spatial structure, geometric consistency, and imaging validity. The inferred degradation evidence is then translated into a score adjustment that corrects the base prediction, while simultaneously providing interpretable degradation perception. In this way, the proposed framework decomposes quality prediction into base estimation and reasoning-guided refinement, enabling multimodal reasoning to complement, rather than replace, learned quality prediction. Such a design effectively bridges the gap between image-level perceptual assessment and scene-level spatial understanding, leading to more accurate and interpretable quality evaluation for 3DGS scenes.

The main contributions of this work are summarized as follows. 

\begin{itemize}
\item To the best of our knowledge, this is the first work to explicitly model 3DGS scene quality through spatially-aware representations, moving beyond conventional image-based IQA. A 3D-aware quality learning paradigm is proposed by equipping a VGGT-based encoder with a dedicated quality head, enabling the modeling of cross-view consistency and geometric structure for scene-level quality prediction.

\item A grounded multimodal reasoning module is introduced to incorporate MLLMs into 3DGS quality assessment in a structured manner. Instead of directly generating quality scores, the MLLM is constrained to perform evidence-based degradation diagnosis using multimodal spatial cues, thereby improving reliability and preventing hallucinated or inconsistent predictions.

\item A reasoning quality prediction paradigm is proposed to reformulate 3DGS quality assessment as base estimation plus reasoning-guided refinement. The reasoning output is translated into a bounded score adjustment, enabling explicit correction of systematic biases while providing interpretable degradation attribution.

\item Extensive experiments on benchmark datasets demonstrate that the proposed method consistently outperforms state-of-the-art IQA-based and MLLM-based approaches, while producing degradation analysis that is well aligned with the underlying spatial characteristics of 3DGS scenes.

\end{itemize}

\section{Related Work}

\subsection{3D guassian splatting compression}

3D Gaussian Splatting has emerged as an effective representation for real-time novel view synthesis. However, the large storage and memory overhead of dense Gaussian representations has stimulated increasing interest in 3DGS compression~\cite{wu2025survey}. LightGS~\cite{fan2024lightgaussian} compresses 3DGS by pruning less important Gaussians and quantizing Gaussian attributes, thereby reducing redundancy in both geometry and appearance parameters. While effective for storage reduction, this strategy often introduces structural sparsity and color discretization artifacts, which may degrade both geometric completeness and visual smoothness. CompGS~\cite{navaneet2024compgs} follows a similar compression principle by combining compact parameter representation with quantization of Gaussian attributes. Such compression typically reduces fine-grained appearance fidelity and may lead to noticeable degradation in local structural detail. C3DGS~\cite{niedermayr2024compressed} further improves compactness through parameter clustering and importance-aware filtering, allowing multiple Gaussians to share representative attributes. Although this design increases compression efficiency, it can also produce prototype-like geometry, reduced local contrast, and weakened structural fidelity in complex regions. Compact 3DGS~\cite{lee2024compact} adopts a more aggressive compression strategy by replacing explicit color representations with hash-grid neural networks and applying multi-level quantization to geometric parameters. This design achieves high compression ratios, but often causes over-smoothing and loss of high-frequency details, which directly affect perceptual sharpness. HAC~\cite{chen2024hac} compresses 3DGS by introducing hierarchical spatial anchors to guide Gaussian generation and quantization. By restructuring the underlying representation, HAC may alter the spatial distribution of Gaussian primitives, which often leads to geometry-dominated degradations such as structural deformation and cross-view inconsistency. Scaffold-GS~\cite{lu2024scaffold} similarly relies on anchor-based representation design, where Gaussian primitives are generated or organized under structural scaffolds. Although this improves representation efficiency, the modified structural organization may introduce reconstruction instability and view-dependent inconsistency, especially in geometrically challenging regions.

Unlike conventional image compression, which mainly introduces pixel-level and largely view-invariant artifacts, 3DGS compression produces representation-level distortions that are inherently geometry-aware and often vary across viewpoints. As a result, the perceptual quality of compressed 3DGS scenes depends not only on appearance fidelity, but also on structural integrity and cross-view coherence. These properties make compressed 3DGS quality assessment different from traditional image quality assessment, and necessitate dedicated quality models that explicitly account for scene-level spatial degradation.

\subsection{Image quality assessment}

Existing image quality assessment (IQA) methods can be broadly categorized into handcrafted metrics and learning-based models. Handcrafted metrics, such as PSNR, SSIM~\cite{wang2004image}, VIF~\cite{sheikh2006image}, and BRISQUE~\cite{mittal2012no}, measure pixel-level fidelity, structural similarity, or statistical regularity in 2D images. Although efficient and widely used, these methods assume view-invariant distortions and are unable to capture geometry-induced and view-dependent artifacts in 3DGS. 

Deep learning–based IQA models, including LPIPS~\cite{zhang2018unreasonable}, DBCNN~\cite{zhang2018blind}, MUSIQ~\cite{ke2021musiq}, and LIQE~\cite{zhang2023blind}, improve perceptual quality prediction by learning feature representations from large-scale datasets. However, these models are primarily trained on conventional 2D distortions and operate on single-view inputs, making them insufficient for modeling scene-level spatial structure and cross-view consistency in 3DGS.

More recently, vision-language and multimodal large models have been introduced for quality assessment. CLIP-IQA~\cite{wang2022exploring} leverages pretrained vision-language representations for text-guided quality prediction, while Q-Align~\cite{wu2024q} improves multimodal alignment for quality reasoning, and Q-Insight~\cite{li2025qinsight} introduces explicit reasoning for joint score prediction and degradation explanation. Despite their strong semantic understanding, these methods rely on image-level representations and lack explicit spatial grounding. Moreover, they are not optimized for stable quantitative regression, limiting their reliability for 3DGS quality prediction.

\subsection{3D quality assessment}

Quality assessment for 3D representations has been explored in both point cloud and neural rendering domains. For point cloud quality assessment (PCQA), MM-PCQA~\cite{zhang2023mm} models multi-modal features to jointly capture geometric and perceptual distortions, COM-PCQA~\cite{11456849} leverages complex-valued representations to encode amplitude and phase information for enhanced structural sensitivity, and MOD-PCQA~\cite{wang2024zoom} introduces multi-order distortion modeling to better characterize degradation patterns in point clouds. While effective for static 3D geometry, these methods do not consider view-dependent rendering or radiance variations, and thus cannot fully reflect the perceptual quality of 3DGS scenes.

In neural rendering, NeRF-QA~\cite{martin2023nerf} evaluates rendering fidelity by measuring perceptual differences across synthesized views, while GS-QA~\cite{martin2025gs} extends quality assessment to Gaussian-based representations by comparing NeRF and 3DGS outputs. GSC-QA~\cite{yang2024benchmark} further investigates the perceptual impact of compression artifacts in 3DGS by analyzing distortion patterns introduced during Gaussian parameter quantization.  Although these approaches incorporate multi-view information, they are either designed for implicit representations or limited to specific distortion settings, and do not fully capture the joint effects of geometry, appearance, and cross-view consistency in compressed 3DGS scenes.

\subsection{MLLMs for spatial visual understanding}

Multimodal large language models (MLLMs) have been increasingly applied to visual understanding and quality-related tasks due to their strong cross-modal reasoning capabilities. LLaVA-one-vision~\cite{xiong2024llavaovchat} (0.5B and 7B) extends large language models with visual encoders to enable image-based reasoning and question answering. However, for quality assessment, it primarily relies on high-level semantic cues and lacks explicit modeling of geometric consistency, often resulting in unstable or subjective predictions. DeepSeekVL~\cite{lu2024deepseek} integrates vision-language pretraining with instruction tuning to improve multimodal reasoning. While effective for general visual understanding, it does not explicitly incorporate spatial structure or multi-view constraints, limiting its capability in geometry-aware quality evaluation. Qwen2.5-VL~\cite{bai2025qwen2} improves multimodal reasoning through large-scale instruction tuning and enhanced visual-language alignment. Despite strong generalization ability, its predictions are sensitive to prompt design and may not consistently align with perceptual quality criteria in 3DGS scenarios. Gemini1.5-pro~\cite{gemini15} demonstrates powerful multimodal reasoning with long-context understanding. However, its quality evaluation is largely semantic and descriptive, without explicit grounding in geometric or structural evidence, which is critical for 3DGS quality assessment. mPLUG-Owl3~\cite{yemplug} enhances multimodal alignment and instruction-following through unified vision-language modeling. Nevertheless, it does not explicitly model 3D spatial cues such as depth consistency or point-cloud structure, making it less suitable for scene-level quality evaluation. InternVL3~\cite{chen2024expanding} scales multimodal pretraining to achieve strong performance across diverse vision-language tasks. However, it treats images as independent inputs and does not explicitly reason over spatial structure or multi-view consistency, which limits its reliability for 3DGS quality assessment. Overall, although these MLLMs exhibit strong capabilities in general visual reasoning, they lack explicit modeling of spatial structure, geometric consistency, and cross-view coherence, which are essential for reliable 3DGS quality assessment.

\section{Method}

\begin{figure*}[t]
\centering
\includegraphics[width=\textwidth]{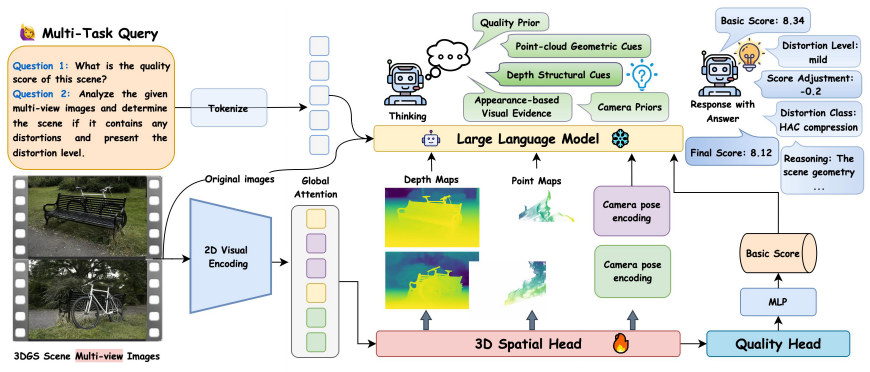}
\caption{Overview of the proposed SpatialQ framework. A spatial quality encoder first processes multi-view rendered images to produce a base quality score and structured geometric representations. The multimodal reasoning model then takes the original images together with encoder-derived depth maps, point-cloud visualizations, and camera parameters as input, and performs degradation diagnosis and score refinement through grounded reasoning. The final quality score is obtained by combining the base prediction with a reasoning-guided adjustment.}
\label{fig2}
\end{figure*}

In this section, the proposed SpatialQ framework as shown in Figure~\ref{fig2} is presented for 3DGS scene quality assessment. The framework aims to unify 3D-aware quality prediction and multimodal reasoning-guided refinement within a single pipeline. It addresses two key challenges: learning scene-level spatial quality representations beyond conventional image-based assessment, and incorporating MLLM reasoning for stable and interpretable score refinement. The following subsections describe the spatial quality encoder, the grounded multimodal reasoning module, and the reasoning-guided refinement strategy.

\subsection{3D-aware quality representation learning}

Given a pair of rendered views from the same 3D reconstruction, quality assessment is formulated as a paired-view representation learning problem rather than independent single-image regression. Let $\mathbf{I}_1$ and $\mathbf{I}_2$ denote two rendered images of the same scene under different viewpoints. Since distortions in 3DGS-based reconstruction are often manifested not only as local appearance degradation but also as cross-view structural inconsistency, the two views are jointly modeled as a single quality sample. The base quality score is defined as
\begin{equation}
Q_{\text{base}} = f(\mathbf{I}_1, \mathbf{I}_2),
\end{equation}
where $f(\cdot)$ captures both perceptual fidelity and structural consistency across viewpoints.

To learn such a representation, a shared 3D-aware visual backbone is employed to encode the paired inputs. Specifically, the two rendered views are processed by a weight-sharing encoder:
\begin{equation}
\mathbf{z}_1 = E(\mathbf{I}_1), \quad \mathbf{z}_2 = E(\mathbf{I}_2),
\end{equation}
where $E(\cdot)$ denotes the feature extraction network. This design ensures consistent representation across viewpoints while preserving view-specific visual characteristics.

To further model cross-view structural relationships, the feature embeddings are integrated through a cross-view interaction function:
\begin{equation}
\mathbf{z}_{\text{joint}} = \mathcal{F}(\mathbf{z}_1, \mathbf{z}_2),
\end{equation}
where $\mathcal{F}(\cdot)$ denotes feature aggregation, implemented via concatenation followed by attention-based fusion. This interaction explicitly encodes both correspondence and discrepancy between views, enabling the model to capture shared scene structures as well as view-dependent variations.

The resulting representation simultaneously captures intra-view perceptual cues and inter-view structural dependencies. Intra-view cues include appearance degradations such as blur, color distortion, and local artifacts, while inter-view relationships reflect geometric consistency and structural coherence under viewpoint changes. In particular, discrepancies between $\mathbf{z}_1$ and $\mathbf{z}_2$ serve as implicit indicators of geometry-related degradation, such as structural misalignment or instability across views. This implicit modeling allows the network to incorporate geometry-aware information without requiring explicit 3D supervision.

Based on the learned representation, a dedicated quality head is introduced to predict view-wise quality scores:
\begin{equation}
s_1 = g(\mathbf{z}_1), \quad s_2 = g(\mathbf{z}_2),
\end{equation}
where $g(\cdot)$ denotes the quality prediction function. The final base quality score is obtained by aggregating the view-wise predictions:
\begin{equation}
Q_{\text{base}} = \phi(s_1, s_2),
\end{equation}
where $\phi(\cdot)$ represents an aggregation function. The view-wise scores preserve local quality variations across viewpoints, while the aggregated score summarizes the overall perceptual and geometric reliability of the reconstructed scene. This base prediction provides a stable data-driven estimate, which is further refined by the subsequent reasoning module using structured spatial evidence.

\subsection{Grounded multimodal reasoning with structured spatial evidence}

To further capture scene-level degradations that are not fully reflected in appearance-based representations, a grounded multimodal reasoning module is introduced to analyze structured spatial evidence. Unlike conventional visual encoders that rely solely on image features, this module explicitly incorporates complementary modalities, including rendered images, depth maps, point cloud visualizations, and camera parameters, to enable comprehensive scene understanding.

Given a reconstructed scene, the multimodal inputs are organized as
\begin{equation}
\mathcal{X} = \{\mathbf{I}, \mathbf{D}, \mathbf{P}, \mathbf{C}\},
\end{equation}
where $\mathbf{I}$ denotes rendered images, $\mathbf{D}$ represents depth maps, $\mathbf{P}$ corresponds to point cloud projections, and $\mathbf{C}$ encodes camera parameters such as intrinsic and extrinsic configurations. These modalities provide complementary information: $\mathbf{I}$ reflects perceptual appearance, $\mathbf{D}$ encodes geometric continuity, $\mathbf{P}$ describes spatial distribution, and $\mathbf{C}$ characterizes viewpoint configuration.

A multimodal large language model (MLLM) is employed to perform structured reasoning over $\mathcal{X}$:
\begin{equation}
\mathcal{R} = f_{\text{MLLM}}(\mathcal{X}),
\end{equation}
where $\mathcal{R}$ denotes the inferred degradation description. Instead of directly predicting quality scores, the MLLM is designed to produce structured outputs that characterize the degradation in terms of type and severity:
\begin{equation}
\mathcal{R} = \{t, s\},
\end{equation}
where $t$ represents the distortion type, and $s$ denotes its severity level.

\begin{table*}[t]
\caption{Shared System Prompt. The system prompt is shared across all reasoning tasks to define a consistent reasoning role, diagnostic scope, evidence source, and score adjustment principle.}
\label{tab:system_prompt}
\centering
\renewcommand{\arraystretch}{1.15}
\setlength{\tabcolsep}{6pt}
\begin{tabular}{p{0.96\textwidth}}
\toprule

\textbf{System Prompt:} A conversation between User and Assistant for 3D scene quality diagnosis. The user provides the original multi-view images and asks a question about scene quality. An quality assessment encoder first processes the images and produces auxiliary structural evidence, including depth maps, point-cloud visualizations, and camera parameters. The assistant then analyzes the original images together with the encoder outputs and the user question to diagnose the quality of the reconstructed 3DGS scene, determine whether the base quality score should be adjusted, and provide an evidence-grounded explanation. The assistant should focus on quality-related evidence rather than generic scene description, and should mention semantic objects only when they are directly relevant to explaining quality degradation. The reasoning process and answer are enclosed within \texttt{<think>} \texttt{</think>} and \texttt{<answer>} \texttt{</answer>} tags, respectively, i.e., \texttt{<think>} reasoning process here \texttt{</think><answer>} answer here \texttt{</answer>}.\\

\midrule

\textbf{Available Evidence:} For each view, you are provided with the rendered image, the corresponding depth map, the point-cloud rendering, and structured statistics. In addition, heuristic textual findings (\texttt{heuristic\_findings}) are provided as auxiliary evidence. All judgments must be grounded in these multimodal inputs. \\

\midrule

\textbf{Diagnostic Dimensions:} Assess the scene by comparing evidence along the following dimensions: color fidelity in the rendered image; geometric deformation in the rendered image; visible artifacts in the rendered image; dynamic range and continuity of the depth map; sparsity, continuity, and structural consistency of the point cloud; plausibility of camera pose and field of view; and whether collapse is present. \\

\midrule

\textbf{Degradation Labels:} The distortion category must be selected from the following closed set: \textit{LightGS}, \textit{Scaffold-GS}, \textit{HAC}, \textit{C3DGS}, \textit{CompGS}, \textit{Compact 3DGS}, or \textit{Unknown}. Do not introduce labels outside this set. \\

\midrule

\textbf{Score Adjustment Principle:} The given base quality score should be actively revised according to the multimodal evidence. If clear degradation evidence is observed, the score adjustment should be negative. If the base score is judged to underestimate the actual quality, the adjustment should be positive. The adjustment should not be trivially set to zero. The magnitude of the adjustment should reflect the severity of the observed problem, with mild, moderate, and severe degradations corresponding to increasingly larger corrections. \\

\bottomrule
\end{tabular}
\end{table*}

By jointly reasoning over multiple modalities, the MLLM is able to associate observed artifacts with their underlying causes. For example, inconsistencies in depth maps indicate geometric instability, irregular point distributions suggest sparse reconstruction, and mismatches across viewpoints imply structural incoherence. This evidence-driven reasoning process ensures that the inferred degradation is grounded in observable spatial cues, rather than unconstrained semantic generation.

The structured degradation output $\mathcal{R}$ is subsequently mapped to a score adjustment term:
\begin{equation}
\Delta Q = h(\mathcal{R}),
\end{equation}
where $h(\cdot)$ denotes a mapping function that translates degradation type and severity into a quantitative correction. This design constrains the role of the MLLM to reasoning and diagnosis, while delegating numerical prediction to a controlled refinement process. As a result, the model avoids unstable or hallucinated predictions and produces interpretable degradation-aware corrections. This design transforms quality assessment from direct regression to reasoning-guided correction.

\subsection{Reasoning-guided quality refinement}

To improve reasoning reliability and output controllability, the prompting strategy is organized into two levels: a shared system prompt and task-specific prompts. The shared system prompt defines a consistent reasoning role, available evidence, diagnostic dimensions, degradation labels and score adjustment principle for all tasks. Specifically, the multimodal model is explicitly instructed to act as a 3D scene quality diagnostician rather than a generic scene description model. Its objective is to assess and refine the quality prediction based on multimodal evidence, rather than describing semantic scene content unless such content is directly relevant to the explanation of quality degradation.

In addition, the system prompt constrains the reasoning process to a set of predefined diagnostic dimensions, including color fidelity, geometric distortion, visible artifacts, depth continuity, point-cloud sparsity and structural consistency, camera validity, and potential scene collapse. The degradation category is further restricted to a closed set of candidate labels, which improves output consistency and reduces uncontrolled generation. On top of these shared instructions, task-specific prompts are introduced for quality refinement and diagnostic explanation, respectively. The former guides the model to determine whether the base score should be increased or decreased according to the observed evidence, while the latter focuses on summarizing the dominant degradation pattern in an interpretable and evidence-grounded manner. The shared system prompt and the task-specific prompts are summarized in Tables~\ref{tab:system_prompt} and~\ref{tab:task_prompt}.

\begin{table*}[t]
\caption{Task-Specific Prompts. Task-specific prompts are further designed for numerical refinement and diagnostic explanation, respectively.}
\label{tab:task_prompt}
\centering
\renewcommand{\arraystretch}{1.15}
\setlength{\tabcolsep}{6pt}
\begin{tabular}{p{0.96\textwidth}}
\toprule

\textbf{Prompt for Quality Refinement Task:} Given the base quality score and the multimodal evidence of a reconstructed 3DGS scene, determine whether the current score is overestimated or underestimated. Diagnose the dominant degradation type from the predefined closed set, assess its severity, and explain the decision using appearance, depth, point-cloud, and camera-related evidence. Then output a bounded score adjustment together with the refined score. The adjustment should follow the score adjustment principle defined in the shared system prompt and must be supported by explicit multimodal evidence. \\

\midrule

\textbf{Prompt for Diagnostic Explanation Task:} Given the refined quality score and the same multimodal evidence, summarize the dominant degradation pattern and explain why the refined score is reasonable. Focus on evidence-grounded interpretation rather than score correction. The explanation should identify the main degradation type, describe its severity, and discuss its impact on overall scene quality using appearance and geometric cues. \\

\bottomrule
\end{tabular}
\end{table*}

This two-level prompt design separates common reasoning principles from task-dependent objectives, thereby improving functional specialization and output consistency.

To integrate multimodal reasoning into the quality prediction process, a decomposition-based formulation is adopted, where the final quality score is expressed as the combination of a base estimate and a reasoning-guided correction:
\begin{equation}
Q_{\text{final}} = Q_{\text{base}} + \Delta Q,
\end{equation}
where $Q_{\text{base}}$ is obtained from the 3D-aware representation learning module (Sec. III-A), and $\Delta Q$ is derived from the multimodal reasoning output (Sec. III-B).

Instead of directly predicting the final score using a single model, the proposed formulation decomposes quality assessment into two complementary components. The base score captures global perceptual quality based on learned visual representations, while the correction term focuses on high-level degradations that are not fully reflected in appearance features, such as geometric inconsistency, structural instability, and cross-view incoherence.

The score adjustment term $\Delta Q$ is computed by mapping the structured reasoning output $\mathcal{R} = \{t, s\}$ into a scalar correction:
\begin{equation}
\Delta Q = h(\mathcal{R}),
\end{equation}
where $t$ denotes the degradation type and $s$ represents its severity. The mapping function $h(\cdot)$ can be implemented as a lightweight regression module or a predefined scoring function that assigns different adjustment magnitudes based on degradation categories and their severity levels. This design ensures that the adjustment is explicitly grounded in interpretable degradation attributes.

To ensure stability and prevent excessive correction, the adjustment term is constrained within a bounded range:
\begin{equation}
\Delta Q \in [-\delta, \delta],
\end{equation}
where $\delta$ is a predefined or learnable threshold. This constraint limits the influence of the reasoning module, ensuring that the final prediction remains consistent with the base estimate while allowing meaningful refinement.

The proposed decomposition offers two key advantages. First, it enables effective correction of systematic biases in the base predictor, particularly in cases where appearance-based representations fail to capture geometry-related degradations. Second, it introduces an interpretable intermediate layer that explicitly links degradation reasoning to quantitative score refinement, thereby transforming quality assessment from a direct regression problem into a reasoning-guided decision process.

The final output $Q_{\text{final}}$ thus reflects both data-driven perception and reasoning-based correction, achieving improved accuracy and interpretability for 3DGS scene quality assessment. This formulation shifts quality assessment from a purely data-driven regression paradigm to a reasoning-aware correction paradigm.

\section{Experiment}

\subsection{Experimental setup}

\textbf{Datasets and metrics.}

For 3DGS quality assessment, we adopt the 3DGS-IEval-15K dataset~\cite{xing20253dgs}, a large-scale benchmark specifically designed for evaluating compressed 3D Gaussian Splatting representations. The dataset contains 15,200 rendered images generated from 10 real-world indoor and outdoor scenes selected from MipNeRF 360~\cite{barron2022mip}, Tanks \& Temples~\cite{knapitsch2017tanks}, and Deep Blending~\cite{hedman2018deep}. These images are produced using six representative 3DGS compression methods under systematically controlled geometry and color compression settings. In total, 760 compressed 3DGS models are constructed, each rendered from 20 viewpoints, including both training views and novel views, to capture view-dependent quality variations. All rendered images are annotated with Mean Opinion Scores (MOS) obtained through controlled subjective studies, resulting in approximately 228,000 human ratings. 

To evaluate the consistency between predicted scores and MOS, we adopt three widely used correlation-based metrics, including Spearman rank-order correlation coefficient (SRCC), Pearson linear correlation coefficient (PLCC), and Kendall rank-order correlation coefficient (KRCC).

The SRCC measures the monotonic relationship between predicted scores and MOS by comparing their rank orders. It is defined as:
\begin{equation}
\mathrm{SRCC} = 1 - \frac{6 \sum_{i=1}^{N} \left( R(x_i) - R(y_i) \right)^2}{N(N^2 - 1)},
\end{equation}
where $x_i$ and $y_i$ denote the predicted score and the corresponding MOS of the $i$-th sample, respectively, and $R(\cdot)$ represents the rank operator.

The PLCC evaluates the linear correlation between predicted scores and MOS, and is computed as:
\begin{equation}
\mathrm{PLCC} =
\frac{\sum_{i=1}^{N} (x_i - \bar{x})(y_i - \bar{y})}
{\sqrt{\sum_{i=1}^{N} (x_i - \bar{x})^2}\sqrt{\sum_{i=1}^{N} (y_i - \bar{y})^2}},
\end{equation}
where $\bar{x}$ and $\bar{y}$ denote the mean values of predicted scores and MOS, respectively.

The KRCC measures the ordinal association between predicted scores and MOS by considering pairwise concordance:
\begin{equation}
\mathrm{KRCC} = \frac{N_c - N_d}{\frac{1}{2}N(N-1)},
\end{equation}
where $N_c$ and $N_d$ denote the numbers of concordant and discordant pairs, respectively.

SRCC and KRCC evaluate the monotonic consistency between predicted scores and MOS, while PLCC measures the linear correlation after regression. These metrics jointly provide a comprehensive evaluation of both ranking capability and prediction accuracy.

\textbf{Implementation details.}

The proposed model is implemented in PyTorch with GPU acceleration. We adopt a VGGT-based backbone~\cite{wang2025vggt} with an additional quality head, and fine-tune all parameters jointly from pretrained weights. For multimodal reasoning, we employ Qwen2.5-VL (7B)~\cite{bai2025qwen2} as the large multimodal language model. The framework supports both single-view and multi-view training through single-image and paired-image supervision. Training is conducted on NVIDIA A800 GPUs with multi-GPU acceleration when available. During inference, an NVIDIA RTX 3090 GPU is used. Mixed-precision training is enabled to improve efficiency. The dataset is randomly split into training and testing sets with a ratio of 8:2. To ensure a fair comparison, all baseline methods are retrained under the same data split protocol. This guarantees that there is no overlap between training and testing samples, thereby avoiding data leakage and enabling unbiased evaluation.

\subsection{Overall performance}

A total of 16 state-of-the-art quality assessment methods are selected for comparison, including handcrafted-based IQA models, deep learning-based IQA models, and MLLM-based zero-shot models. The handcrafted-based IQA models include PSNR, SSIM~\cite{wang2004image}, VIF~\cite{sheikh2006image}, and BRISQUE~\cite{mittal2012no}. These methods are primarily designed for pixel-level distortion measurement and are applied to rendered images for evaluation. The deep learning-based IQA models include LPIPS(VGG)~\cite{zhang2018unreasonable}, DBCNN~\cite{zhang2018blind}, TReS~\cite{golestaneh2022no}, and DISTS~\cite{ding2020image}. These methods learn perceptual features from images and have shown strong performance on image quality assessment tasks. The MLLM-based zero-shot models include LLaVA-one-vision (0.5B and 7B)~\cite{xiong2024llavaovchat, xiong2024llavaovchat}, DeepSeekVL (7B)~\cite{lu2024deepseek}, Qwen2.5-VL (7B)~\cite{bai2025qwen2}, Gemini1.5-pro~\cite{gemini15}, mPLUG-Owl3 (7B)~\cite{yemplug}, Q-Align~\cite{wu2024q}, and InternVL3 (9B)~\cite{chen2024expanding}. These models are designed for general multimodal understanding and are directly applied to quality assessment. All IQA-based methods operate on rendered images without explicit modeling of 3D spatial structure, while MLLM-based methods rely on general visual reasoning without dedicated training for quality regression. The proposed SpatialQ is compared against these methods to evaluate its effectiveness in 3DGS quality assessment.

Table~\ref{tab1} reports the benchmarking results on the 3DGS-IEval-15K~\cite{xing20253dgs} dataset. The proposed SpatialQ achieves the best overall performance across all evaluation settings, demonstrating clear advantages over handcrafted IQA methods, deep learning-based IQA models, and MLLM-based zero-shot approaches.

Compared with handcrafted IQA metrics, SpatialQ shows substantial improvements in both SRCC and PLCC. This is expected, as these traditional metrics are primarily designed for pixel-level distortion and fail to capture geometry-aware degradations and cross-view inconsistencies inherent in compressed 3DGS scenes. 

Compared with deep learning-based IQA models, SpatialQ consistently achieves higher correlation with MOS. While these models learn perceptual features from images, they still operate on single-view inputs and lack explicit modeling of spatial structure and multi-view consistency, which limits their effectiveness for 3DGS quality assessment.

MLLM-based zero-shot methods exhibit competitive performance compared to IQA baselines, indicating their strong visual reasoning capability. However, their performance remains inferior to SpatialQ, particularly in terms of stability and correlation consistency. This is mainly due to the lack of explicit spatial grounding and their sensitivity to prompt design, which makes it difficult to produce reliable quantitative predictions.

\begin{table}[t]
\footnotesize
\caption{Benchmarking results of state-of-the-art methods on the 3DGS-IEval-15K dataset, including three outdoor and three indoor scenes from MipNeRF 360, two outdoor scenes from Tanks \& Temples, and two indoor scenes from Deep Blending. Best results are highlighted in red and the second best in blue.}
\label{tab1}
\def\tabblank{\hspace*{7mm}}
\begin{tabularx}{\textwidth}{Xlcccccccccc}
\toprule
\multirow{2}{*}{Index} & \multirow{2}{*}{Methods/Metric} & \multicolumn{3}{c}{All} & \multicolumn{3}{c}{Geometry-Only} & \multicolumn{3}{c}{Color \& Geometry Mix} \\ 
\cline{3-11}
& & SRCC$\uparrow$ & PLCC$\uparrow$ & KRCC$\downarrow$ & SRCC$\uparrow$ & PLCC$\uparrow$ & KRCC$\downarrow$ & SRCC$\uparrow$ & PLCC$\uparrow$ & KRCC$\downarrow$ \\ \cline{1-11}
&Handcrafted-based IQA Models&\\ \hdashline
A&PSNR & 0.6451 & 0.6387 & 0.4574 & 0.4769 & 0.5141 & 0.3326 & 0.6372 & 0.6493 & 0.4547 \\
B&SSIM~\cite{wang2004image} & 0.6790 & 0.6651 & 0.4889 & 0.6246 & 0.6448 & 0.4475 & 0.6687 & 0.6542 & 0.4844 \\
C&VIF~\cite{sheikh2006image} & 0.5590 & 0.5689 & 0.3928 & 0.5525 & 0.5644 & 0.3907  & 0.5799 & 0.6062 & 0.4108 \\
D&BRISQUE~\cite{mittal2012no} & 0.2202 & 0.2254 & 0.1495 & 0.3746 & 0.3734 & 0.2560 & 0.2594 & 0.2591 & 0.1759 \\ \hline
& IQA Models&\\ \hdashline
E&LPIPS(VGG)~\cite{zhang2018unreasonable} & 0.7426 & 0.7409 & 0.5367 & 0.7173 & 0.7152 & 0.5229 & 0.7458 & 0.7589 & 0.5545 \\
F&DBCNN~\cite{zhang2018blind} & \textcolor{blue}{0.8635} & \textcolor{blue}{0.8551} & 0.6699 & \textcolor{blue}{0.7770} & \textcolor{blue}{0.7896} & 0.5817 & \textcolor{blue}{0.8471} & \textcolor{blue}{0.8536} & 0.6578 \\
G&TReS~\cite{golestaneh2022no}&0.7969 &0.7966 &\textcolor{blue}{0.5988}& 0.6578& 0.6891& \textcolor{blue}{0.4749}&0.7912 &0.8005 &\textcolor{blue}{0.5988}\\
H&Q-Align~\cite{wu2024q} & 0.7711 & 0.7646 & 0.5668 & 0.7377 & 0.7413 & 0.5365 &  0.7554 & 0.7657 & 0.5575 \\
I&DISTS~\cite{ding2020image} & 0.8198 & 0.8132 & 0.6215 & 0.7593 & 0.7699 & 0.5613& 0.8342 & 0.8288 & 0.6428 \\
\hline
&MLLM-based Zero-Shot Models&\\ \hdashline
J&Llava-one-vision (0.5B)~\cite{xiong2024llavaovchat}& 0.2286 & 0.2311 & 0.1867 & 0.1075 & 0.1179 & 0.0876 &  0.2180 & 0.2353 & 0.1783 \\
K&Llava-one-vision (7B)~\cite{xiong2024llavaovchat} & 0.4628 & 0.4821 & 0.3774 & 0.2690 & 0.3323 & 0.2204  & 0.5197 & 0.5048 & 0.4232 \\
L&DeepSeekVL (7B)~\cite{lu2024deepseek} & 0.7065 & 0.6891 & 0.5480 & 0.6693 & 0.6621 & 0.5134  & 0.6778 & 0.6855 & 0.5320 \\
M&Qwen2.5-VL (7B)~\cite{bai2025qwen2} & 0.7275 & 0.6985 & 0.5806 & 0.7114 & 0.6851 & 0.5656 & 0.6992 & 0.6970 & 0.5606 \\
N&Gemini1.5-pro~\cite{gemini15} & 0.6880 & 0.0893 & 0.5410 & 0.7125 & 0.1041 & 0.5589 & 0.6653 & 0.1085 & 0.5309 \\
O&mPLUG-Owl3 (7B)~\cite{yemplug} & 0.3575 & 0.3515 & 0.2687 & 0.5452 & 0.5616 & 0.4291 & 0.2960 & 0.3009 & 0.2234 \\
P&InternVL3 (9B)~\cite{chen2024expanding}&0.5052 & 0.5032 & 0.3796 & 0.5901 &0.5592 & 0.4420& 0.4824& 0.5103& 0.3666\\
\hline
&Proposed Method&\\ \hdashline
 Q&SpatialQ (ours)& \textcolor{red}{0.8937} & \textcolor{red}{0.8840} & \textcolor{red}{0.6161} & \textcolor{red}{0.8099} & \textcolor{red}{0.8073} & \textcolor{red}{0.5080} & \textcolor{red}{0.8835} & \textcolor{red}{0.8804} & \textcolor{red}{0.5879} \\
\bottomrule
\end{tabularx}
\end{table}

Further analysis across different distortion settings shows that SpatialQ maintains strong performance in both geometry-only and color \& geometry mixed scenarios. In the geometry-only subset, the proposed method achieves the highest SRCC and PLCC, demonstrating its ability to capture geometry-induced degradations such as structural inconsistency and depth distortion. In the color \& geometry mixed setting, SpatialQ also achieves the best overall performance, indicating that the proposed framework can jointly model appearance and structural quality factors.

\subsection{Scene reasoning comparison}

\begin{figure*}[htbp]
\centering
\includegraphics[width=\textwidth]{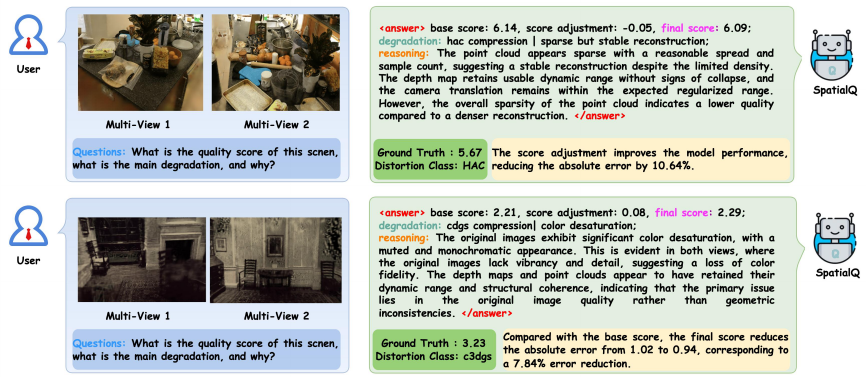}
\caption{Qualitative results of score prediction and explanation by SpatialQ. The model demonstrates the ability to recognize textual content, analyze depth maps and point-cloud projections, and understand scene composition for comprehensive quality assessment.}
\label{fig3}
\end{figure*}

Figure~\ref{fig3} presents qualitative examples illustrating the effectiveness of the proposed framework in refining quality prediction through multimodal reasoning. For each case, the model outputs a base score, a reasoning-guided score adjustment, a final score, and an explicit degradation interpretation.

In the first example, HAC compression, the base score 6.14 slightly overestimates the ground-truth MOS 5.67. Through multimodal reasoning over point cloud sparsity, depth stability, and camera motion, the model identifies the primary degradation as sparse reconstruction under HAC compression. Based on this diagnosis, a negative score adjustment -0.05 is applied, yielding a refined final score 6.09 that is closer to the ground truth. Notably, the reasoning explicitly explains that although the reconstruction remains stable in terms of depth and camera consistency, the reduced point density limits overall quality, demonstrating the model’s ability to balance geometric stability and structural completeness.

In the second example, C3DGS compression, the base score 2.21 underestimates the ground-truth MOS 3.23. The MLLM identifies color desaturation as the dominant degradation while recognizing that geometric structure (depth and point cloud) remains largely intact. Consequently, a positive score adjustment +0.08 is introduced, correcting the bias of the base prediction and producing a more accurate final score 2.29. This case highlights the model’s capability to distinguish between appearance degradation and geometric consistency, avoiding over-penalization when structural integrity is preserved.

Across both examples, the score adjustment consistently reduces the absolute error (by 10.64\% and 7.84\%, respectively), demonstrating that the reasoning-guided refinement effectively corrects systematic biases in the base predictor. More importantly, the generated degradation reasoning aligns well with the observed scene characteristics, providing interpretable evidence for the adjustment process.

Overall, Figure~\ref{fig3} validates that the proposed framework not only improves quantitative prediction accuracy, but also enables fine-grained degradation attribution and explainable score correction, which are essential for reliable 3DGS scene quality assessment.

\subsection{Distortion perception}

\begin{figure*}[htbp]
\centering
\includegraphics[width=\textwidth]{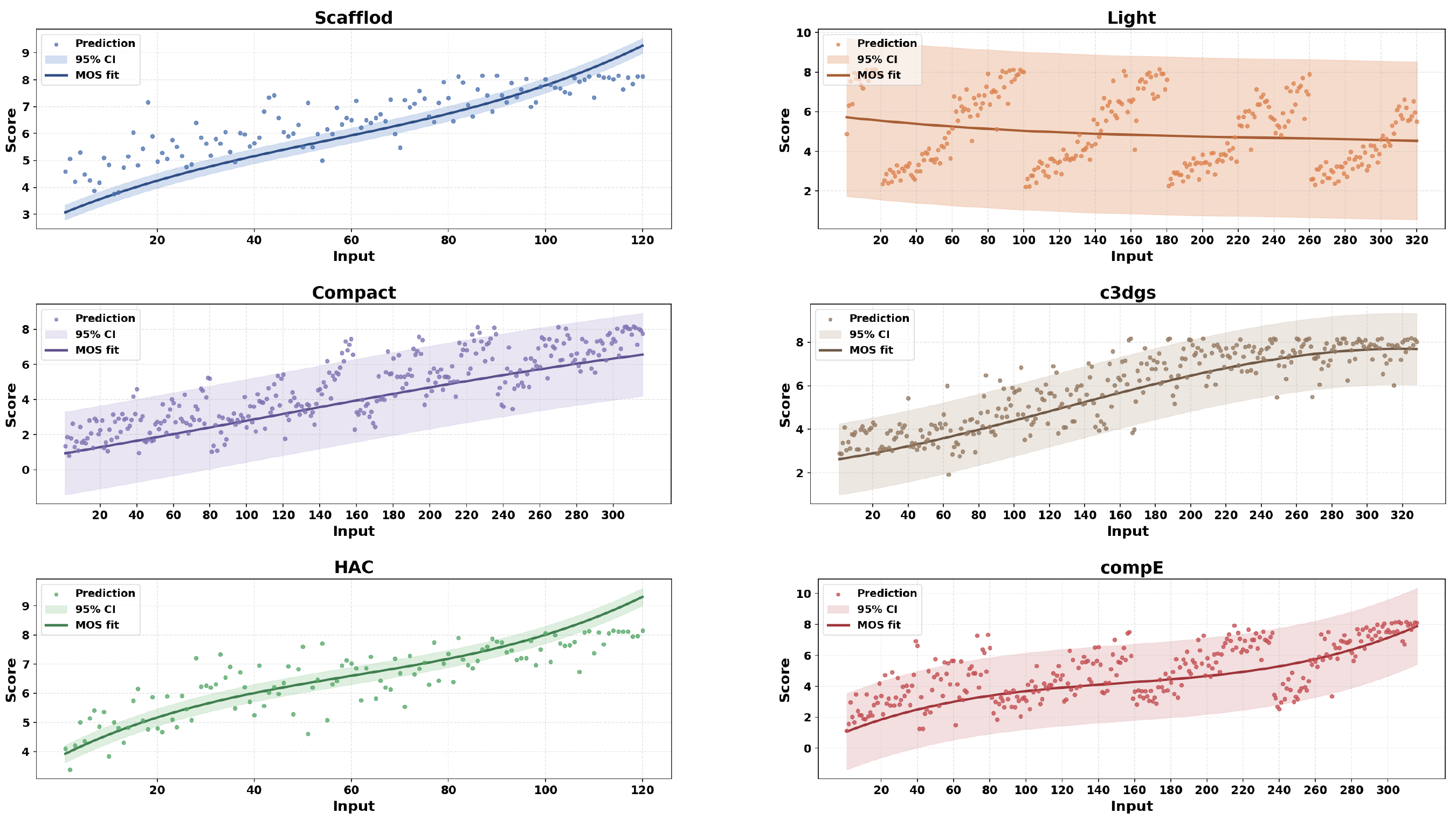}
\caption{Scatter plots of the predicted quality scores versus the MOS scores for different distortion type on the 3DGS-IEval-15K~\cite{xing20253dgs} dataset.}
\label{fig4}
\end{figure*}

Figure~\ref{fig4} illustrates the relationship between predicted quality scores and ground-truth MOS across different 3DGS compression schemes, including Scaffold, Light, Compact, C3DGS, HAC, and CompE. Therefore, the proposed method demonstrates a clear monotonic relationship with MOS in most cases, indicating strong consistency with human perceptual judgments.

From a global perspective, most subfigures (Scaffold, Compact, C3DGS, HAC, and CompE) exhibit a stable positive correlation between predicted scores and MOS. The fitted curves closely follow the distribution of scatter points, suggesting that the model can effectively capture quality variations across different compression levels. In particular, for Scaffold and HAC, the predictions show tight clustering around the fitted curve with relatively narrow confidence intervals, indicating both high prediction accuracy and strong stability. Similarly, in C3DGS, the model maintains consistent alignment with MOS across a wide score range, demonstrating robustness under diverse structural distortions.

For more challenging compression schemes such as Compact and CompE, although the overall trend remains positively correlated, the scatter distribution becomes more dispersed and the confidence intervals widen, especially in mid-quality ranges. This suggests that these compression methods introduce more complex or mixed distortions, increasing the difficulty of precise quality estimation. Nevertheless, the fitted curves still preserve correct monotonicity, indicating that the model maintains reliable ranking capability even under challenging conditions.

A notable exception is the Light compression scheme, where the correlation between predicted scores and MOS is relatively weak, and the fitted curve appears nearly flat. This implies that the perceptual quality variation induced by Light compression is less distinguishable or less aligned with the model’s learned representation. It may also indicate that such distortions are subtle, distributed, or not easily captured by appearance and geometry cues alone, posing a challenge for both data-driven prediction and reasoning-based adjustment.

The results demonstrate that the proposed method achieves strong consistency, robustness, and generalization ability across a variety of 3DGS compression scenarios. The tight alignment between predictions and MOS in most cases validates the effectiveness of combining base score prediction with reasoning-guided refinement, while the observed variations across compression types further highlight the intrinsic complexity of 3DGS quality assessment.

\subsection{Score regression}

Table~\ref{tab:scene_comparison} presents the performance of the proposed SpatialQ across three representative datasets, including MipNeRF 360, Tanks \& Temples, and Deep Blending, covering both indoor and outdoor scenes. The results demonstrate that SpatialQ achieves consistently high performance across all datasets, indicating strong generalization capability under diverse scene conditions.

Specifically, on MipNeRF 360~\cite{barron2022mip}, which contains both indoor and outdoor scenes with rich geometric and appearance variations, SpatialQ achieves the highest performance (SRCC: 0.8971, PLCC: 0.8991, KRCC: 0.7642). This suggests that the proposed model effectively captures both structural consistency and perceptual fidelity in complex real-world scenarios. 

On Tanks \& Temples~\cite{knapitsch2017tanks}, which mainly consists of outdoor scenes with large-scale geometry and sparse viewpoints, the performance slightly decreases (SRCC: 0.8837, PLCC: 0.8875, KRCC: 0.6414). This can be attributed to the increased geometric complexity and potential cross-view inconsistencies, which pose greater challenges for quality prediction. Nevertheless, SpatialQ still maintains competitive performance, demonstrating robustness to geometry-dominated distortions.

On Deep Blending~\cite{hedman2018deep}, which focuses on indoor scenes with relatively stable structure but subtle appearance variations, SpatialQ achieves strong performance (SRCC: 0.8957, PLCC: 0.8931, KRCC: 0.6261). This indicates that the model is also effective in capturing fine-grained perceptual degradations in more controlled environments.

The consistent performance across datasets with different characteristics validates the effectiveness of the proposed spatial representation and reasoning-guided refinement mechanism. The results further demonstrate that SpatialQ is capable of handling both appearance-driven and geometry-driven distortions, and generalizes well across diverse 3DGS reconstruction scenarios.

\begin{table}[t]
\footnotesize
\caption{Performance comparisons on the indoor and outdoor scenes from three different datasets.}
\label{tab:scene_comparison}
\def\tabblank{\hspace*{7mm}}
\centering
\begin{tabularx}{\textwidth}{>{\centering\arraybackslash}p{1cm} p{10cm} p{1.5cm}p{1.5cm}p{1.5cm}}
\toprule
Index & Scene Dataset & SRCC$\uparrow$ & PLCC$\uparrow$& KRCC$\downarrow$ \\
\midrule
A & MipNeRF 360~\cite{barron2022mip} (bicycle, flowers, garden, counter, kitchen, room) & 0.8971 & 0.8991 & 0.7642 \\
B & Tanks \& Temples~\cite{knapitsch2017tanks} (train, truck) & 0.8837 & 0.8875 & 0.6414 \\
C & Deep Blending~\cite{hedman2018deep} (playroom, drjohnson)   & 0.8957 & 0.8931 & 0.6261 \\
\bottomrule
\end{tabularx}
\end{table}

\subsection{Ablation studies}

Table~\ref{tab:ablation_training} presents the ablation study on different training strategies. The results clearly demonstrate that each component of the proposed framework contributes to the final performance, and the full model achieves the best overall results.

Starting from the baseline setting with a frozen feature extractor and a simple regression head, the performance is relatively limited, yielding an SRCC of 0.7639. Replacing the regression head with the proposed quality head leads to a clear improvement, increasing the SRCC to 0.8353. This result indicates that the dedicated quality head is more effective than direct regression in extracting quality-relevant information.

When the feature extractor is further optimized in an end-to-end manner, the performance improves substantially, reaching an SRCC of 0.8935 and a PLCC of 0.8801. This demonstrates the importance of jointly adapting the feature representation to 3DGS quality assessment, and suggests that frozen features are insufficient for capturing complex spatial degradations.

After introducing the MLLM-based reasoning module, the overall performance is further improved, with the best results achieved across all metrics, including an SRCC of 0.8937 and a PLCC of 0.8840. Although the gain in SRCC is relatively modest, the consistent improvement in PLCC indicates that the reasoning-guided refinement helps reduce prediction bias and improves alignment with subjective quality scores.

The ablation results show that the effectiveness of the proposed framework comes from three aspects: the dedicated quality head for improved quality representation learning, end-to-end feature optimization for better spatial degradation modeling, and the reasoning-guided refinement mechanism for more accurate and robust prediction.

\begin{table*}[t]
\footnotesize
\caption{Ablation of the training strategy on 3DGS-IEval-15K~\cite{xing20253dgs} dataset.}
\label{tab:ablation_training}
\centering
\begin{tabularx}{\textwidth}{>{\centering\arraybackslash}p{0.8cm} p{10cm} p{1.5cm}p{1.5cm}p{1.5cm}}
\toprule
Index & Strategy & SRCC$\uparrow$ & PLCC$\uparrow$& KRCC$\downarrow$ \\
\midrule
A & Frozen feature extra heads + regression & 0.7639 & 0.7633 & 0.5661  \\
B & Frozen feature extra heads + quality head & 0.8353 & 0.8318 & 0.6391 \\
C & Training feature extra heads + quality head & 0.8935 & 0.8801 & 0.7135 \\
D & Training feature extra heads + quality head + MLLM & 0.8937 & 0.8840 & 0.6161 \\
\bottomrule
\end{tabularx}
\end{table*}

\section{Conclusion}

This paper presents SpatialQ, a multimodal framework for 3DGS scene quality assessment. By combining 3D-aware quality representation learning with reasoning-guided refinement, the proposed method captures both perceptual fidelity and scene-level spatial degradation. A spatial quality encoder provides a stable base prediction, while a grounded MLLM refines the score through multimodal degradation reasoning. Extensive experiments show that SpatialQ consistently outperforms existing IQA-based and MLLM-based methods across different datasets and distortion settings. In addition to improved prediction accuracy, the proposed framework also provides interpretable degradation analysis. Future work will further explore the role of MLLMs in 3DGS quality assessment, particularly in improving score precision and enhancing deeper understanding of complex spatial degradations, as well as extending the framework to broader 3D scene representations.

%
%
%
%

	\Acknowledgements{This work was supported by The Major Key Project of PCL (PCL2024A02), Natural Science Foundation of China (62271013), Guangdong Provincial Key Laboratory of Ultra High Definition Immersive Media Technology (2024B1212010006), Guangdong Province Pearl River Talent Program (2021QN020708), Guangdong Basic and Applied Basic Research Foundation (2024A1515010155), Shenzhen Science and Technology Program (JCYJ20240813160202004, JCYJ20230807120808017, SYSPG20241211173440004), Shenzhen Fundamental Research Program (GXWD20201231165807007-20200806163656003), and financially supported for Outstanding Talents Training Fund in Shenzhen.}

	

\bibliographystyle{scis}
\bibliography{ref.bib}
    
%
%
%

        
        

	
\end{document}